\documentclass[journal]{IEEEtran}
\usepackage{graphicx}

\ifCLASSINFOpdf
\else
\fi
\begin{document}
\title{Towards Language-Based Modulation of Assistive Robots through Multimodal Models}

\author{Philipp~Wicke,
        L\"{u}fti~Kerem~\c{S}enel,
        Shengqiang~Zhang,
        Luis~Figueredo,
        Abdeldjallil~Naceri,
        Sami~Haddadin
        and~Hinrich~Sch\"{u}tze% <-this % stops a space
\thanks{P. Wicke, L.K. \c{S}enel, S. Zhang and H. Sch\"{u}tze are with 
the Center for Information and Language Processing (CIS) and afflilated 
with the Munich Center for Machine Learning (MCML) at the Ludwig-Maximilians-
University (LMU) Munich, Germany. E-mail: \texttt{\{pwicke, lksenel,  shengqiang\}@cis.lmu.de}
}% <-this % stops a space
\thanks{L. Figueredo, A. Naceri and S. Haddadin are with 
the Munich Institute of Robotics and Machine Intelligence (MIRMI)
at Technical University Munich (TUM), Germany. E-mail:
\texttt{\{luis.figueredo,djallil.naceri,haddadin\}@tum.de}}% <-this % stops a space
}


\markboth{Geriatronics Summit 2023, July 02 - 03,
Garmisch-Partenkirchen Congress Center}%
{Geriatronics Summit 2023}



\maketitle

\begin{abstract}
  In the field of Geriatronics, enabling effective and transparent communication between humans 
  and robots is crucial for enhancing the acceptance and performance of assistive robots. 
  Our early-stage research project investigates the potential of language-based modulation as a 
  means to improve human-robot interaction. We propose to explore 
  real-time modulation during task execution, leveraging language cues, visual 
  references, and multimodal inputs. By developing transparent and interpretable 
  methods, we aim to enable robots to adapt and respond to language commands, enhancing 
  their usability and flexibility. Through the exchange of insights and knowledge at 
  the workshop, we seek to gather valuable feedback to advance our research and 
  contribute to the development of interactive robotic systems for Geriatronics 
  and beyond.
\end{abstract}

\begin{IEEEkeywords}
  HRI, Geriatronics, NLP, Task Modulation
\end{IEEEkeywords}

% \fi
%\IEEEpeerreviewmaketitle

\section{Introduction}
In the field of Geriatronics, where robots play a crucial role in assisting 
older adults with daily tasks, effective human-robot interaction is paramount. 
Imagine a scenario where an older adult needs assistance with taking medication. 
The robot is instructed to ``bring the pill bottle from the kitchen counter.'' 
However, what if the user realizes they need to clarify their request by adding, 
``not the red, but the white bottle''? This language-based modulation is a fundamental 
aspect of language-mediated robot control that can significantly enhance the usability 
and adaptability of robots in real-world scenarios.

Addressing the limitations of language-mediated robot control and the need for 
modulation, this research project aims to explore and develop methods for effective 
modulation in human-robot interaction. Its first focus is on creating a dataset for supporting 
machine learning research on modulation. To achieve this, a diverse dataset will be constructed 
through crowdsourcing, incorporating a range of household tasks relevant to Geriatronics. 
Instructions for modulations will be collected, encompassing both low-level modifications 
that affect specific parameters during task execution and high-level modifications that 
involve adding, removing, or modifying tasks within a larger context. The dataset will 
include instructions with varying levels of specificity to enable the training of 
sophisticated data-driven models. In a second step, the focus shifts towards the design
of transparent methods that leverage deep learning techniques with multimodal models
while providing explanations for the robot's behavior. 

This early-stage research project presents the foundation 
required to complete the envisioned scenarios of language-based modulation
in the context of Geriatronics and beyond.

\section{Related Works and Background}

We provide a comprehensive overview of the research landscape related to
 language-mediated modulation in the context of human-robot interaction. 
 This section is divided into two parts to address distinct aspects of the topic. 
 The first part focuses on the need for a benchmark dataset specifically designed 
 for language-mediated modulation tasks, highlighting the limitations of 
 existing datasets. The second part introduces a transparent approach that 
 incorporates real-time segmentation to enhance the understanding of robot 
 actions during modulated tasks and the robot's task performance itself.

\paragraph*{Language-Mediated Modulation}
In the field of human-robot communication, the ability to modulate robot actions 
using natural language is crucial. However, there is currently a lack of rich 
datasets that specifically address language-mediated modulation. To address this 
gap, our project focuses on creating a benchmark dataset for language-mediated 
modulation and sharing it with the research community. This dataset will enable 
the development of multimodal machine learning approaches for improving human-robot 
interaction.

Existing datasets in the field of robot learning and language-conditioned policies 
provide valuable resources for training models to perform various household tasks. 
On the one hand, datasets such as LangLFP \cite{lynch2021langlfp}, LOREL \cite{nair2022lorel}, 
BC-Z \cite{jang2022bcz}, SayCan \cite{brohan2022saycan}, RT-1 \cite{brohan2022rt1}, 
language-table \cite{lynch2022language} or Latte \cite{bucker2022latte} consist of 
labeled trajectories with language annotations, enabling the learning of 
language-conditioned policies for robot control in performing various low-level 
tasks, such as ``navigate to the kitchen'', ``place the book on the table'', or ``avoid 
obstacles on the floor''.

On the other hand, datasets such as ALFRED \cite{shridhar2020alfred}, 
DialFRED \cite{gao2022dialfred}, Playhouse \cite{team2021play}, TEACh 
\cite{padmakumar2022teach}, and TEAChDA \cite{gella2022teachda} are oriented 
towards accomplishing complex objectives at a higher level, involving the 
learning of action plans with multiple substeps, such as ``prepare a meal'' or
``assemble furniture''. They encompass multiple modalities, integrating visual 
and language information. The degree of interaction between humans and robots
varies among these datasets, with some primarily focusing on the initial 
instruction without extensive ongoing interaction. However, most of these datasets 
do not consider language-mediated modulation specifically, and their focus is
either on single-step tasks or multi-step action plans without explicit 
interaction or modulation during execution.

\paragraph*{Transparent Approach Using Real-Time Segmentation}

\begin{figure*}[t!]
    \centering
    \includegraphics[width=\textwidth]{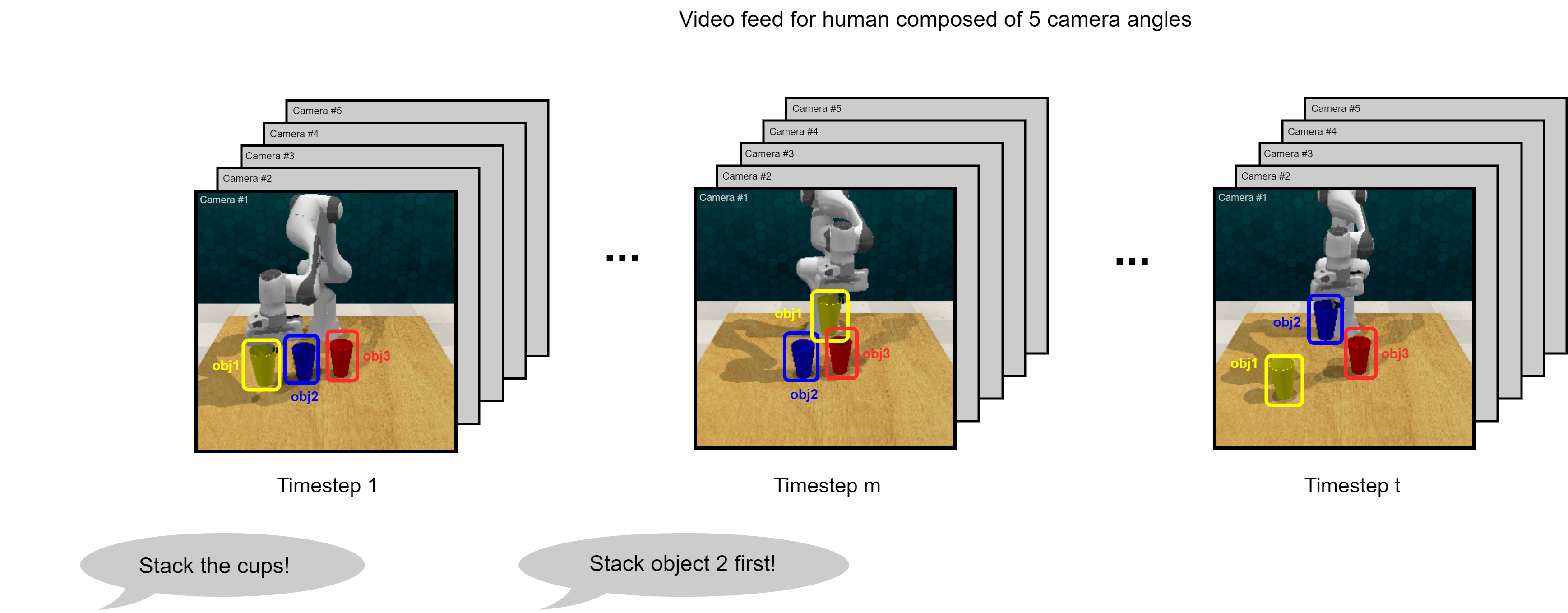}
    \caption{Demonstration of the transparent approach
    for language-based modulation. The initial task
    \textit{stack the cups} is modulated with the command
    \textit{stack object 2 first!} These
    object labels are available to the user in real-time 
    throughout the task execution. For the underlying model, 
    segmentation and label of the target object become available upon 
    modulation for all different camera perspectives.
    \label{fig:demo}}
\end{figure*}

The second part of the proposed research focusses on a
transparent approach for language-based modulation that enables 
robust reference to environmental elements in human-robot communication.
In this context, modulation involves modifying previous 
commands or ongoing actions, requiring the robot to establish 
references to facilitate effective interaction with (elderly) individuals. 
To address this, we aim to create a shared scene representation that both 
humans and robots can access and interpret, facilitating accurate 
references and enhancing communication and interaction for both sides.

Related works have explored the concept of transparency in 
language-mediated robot control. For instance, researchers 
have demonstrated the utility of object labeling in facilitating 
explicit learning or differentiation of novel objects 
\cite{hirschmanner2021investigating,schuetz2017spoken,tucker2020learning}. 
In the context of Geriatronics, this could involve labeling and 
recognizing various everyday objects used by elderly individuals, 
such as medication bottles, walkers, or assistive devices. 
Additionally, transparent methods have been employed to enhance 
communication between robots and humans in various settings 
\cite{bucker2022reshaping,liu2017ar,wicke2021are}. 
For example, a spatial referencing system for mobile robots allows users 
to add waypoints using augmented reality, enabling precise navigation 
assistance \cite{chacko2020augmented}. 
Similarly, in an industrial robot context, visual tracking and 
automated labeling of objects ensure transparent communication through shared labels \cite{liu2017ar}.

Furthermore, human-robot dialogue has been explored as a means to clarify 
ambiguous references and improve language understanding 
\cite{vogel2010eye,parde2015grounding,thomason2016learning,yang2018visual}. 
These studies focus on learning to ground object attributes and names through 
dialogue interactions, which could be applicable to Geriatronics scenarios 
where the robot needs to understand and interpret specific instructions 
related to elderly individuals' needs and preferences. Additionally, 
interactive systems that allow robots to ask follow-up clarifications to 
disambiguate described objects have shown improvements in correctly identifying 
objects with fewer attempts \cite{dogan2022asking}. This approach could be beneficial in situations 
where a person provides instructions to a robot and the robot seeks 
clarification to ensure accurate actions.

Our proposed approach to transparency in language-based modulation differs 
from previous works in two core aspects. Firstly, the language-based interaction 
directly modulates the robot's ongoing actions, aiming to provide immediate 
assistance to individuals. Secondly, our shared scene 
representation serves dual purposes: updating humans about the robot's 
understanding of the environment and supporting natural and robust modulation 
commands. This allows references in natural language commands, such as 
specifying a particular medication or indicating the location of an object,
to be directly linked to objects in the shared scene representation. 
Given that misunderstandings may be critical, this transparency can improve
the user's trust in the system and the interaction. 

\section{Proposed Framework}

\paragraph*{Robotic Environment}

We will utilize RLBench \cite{james2020rlbench} to build our 
dataset by curating and augmenting tasks from its task library.
RLBench offers a convenient platform for dataset creation due to 
its accessibility and task customization capabilities. 
Specifically, it provides automatically generated demonstrations
for a diverse set of one hundred household tasks, which we can 
augment with modulations for our dataset.

RLBench supports multiple variations of tasks, allowing us to 
introduce variability in factors such as target objects and 
colors. Moreover, it enables the generation of unlimited 
episodes for each task variation, enabling us to vary object 
positions across episodes. By incorporating high and low-level 
modulations into the RLBench tasks, we can further diversify 
the dataset. Modulation points can be strategically introduced 
at different stages of task execution, capturing a range 
of modulation scenarios, adding further complexity to the
dataset.

Using RLBench as our dataset generation environment offers 
several advantages. Firstly, RLBench provides a comprehensive 
task library that covers various household tasks, making it 
suitable for studying language-based modulation in the context 
of human-robot interaction. Secondly, the availability of 
automatically generated demonstrations saves time and effort 
in dataset creation. Lastly, the flexibility of RLBench allows 
for easy customization of tasks and the introduction of 
modulations, facilitating the creation of a diverse 
dataset.

\subsection{Dataset Collection and Curation}

Our dataset will incorporate paired demonstrations of low- 
and high-level modulations accompanied by natural language 
instructions, with a possible focus on applications in Geriatronics. 
To gather these instructions, we will employ crowdsourcing, aiming 
to collect approximately 50 natural language instructions for each 
modulation in 30 different household-related tasks. Four conditions
will be considered, resulting from two levels of modulation 
(low- and high-level) and two specificity levels of instructions
(low- and high-specificity).

Low-level (LL) modulations involve modifying a specific parameter 
of an action as it occurs in the Geriatronics context. For instance, 
the instruction ``hand me the pill bottle'' can be LL-modulated by 
saying ``not the brown, but the white one'' or ``be gentle to move it.''
On the other hand, high-level (HL) modulations involve specifying, 
modifying, adding, or removing low-level tasks within the context 
of a high-level task. HL modulations serve to address vagueness in 
the wording of the modulated command or to modify the command itself. 
For example, the command ``bring the cane'' may require the 
modulation command ``avoid stepping on the carpet'' when the robot 
starts walking across the room.

Two types of instructions will be collected: high-specificity 
(HS) and low-specificity (LS). HS modulations provide a detailed 
description of the desired effect of the modulation, leaving 
little ambiguity. Conversely, LS modulations are brief, vague, 
or ambiguous modulations, relying mostly on the information 
provided by the high-level task instruction. 

Crowdsourcing involves presenting crowdworkers with two videos: 
one depicting the robot performing the task without modulation 
and another showing the task with a visual cue indicating modulation.
The crowdworkers are then asked to provide two commands that 
could trigger the modulation: a brief and natural command (LS) 
and a precise and detailed command (HS). Importantly, 
crowdworkers only observe the effect of modulation on 
the robot's behavior and do not receive a modulation 
command directly, as they provide the command.

To facilitate crowdsourcing, the videos include a red exclamation 
mark as a ``modulation point,'' indicating the moment where the 
robot's behavior deviates from its original execution. 
Multiple exclamation marks may be used for tasks with multiple 
modulations. Crowdworkers are requested to rate the difficulty of 
generating the appropriate command and provide an explanation if 
they were unable to generate a command.

For each modulation instance corresponding to a unique modulation 
video in the Geriatronics domain, we will collect a minimum of 30 
high-specificity and low-specificity modulation commands. Leveraging
the capabilities of RLBench tasks, we can create multiple modulation
instances, ensuring the generation of many, diverse data points.

\subsection{Transparent Approach for Language-Based Modulation}

Automatic computation of accurate and semantically rich object
labels is one of the long-term goals of vision research. 
While much progress has been made, the state of the art of 
this technology (e.g., YOLO \cite{wang2022yolo}, DETR 
\cite{carion2020detr}) is not always robust. 
More importantly, even if a semantic label is correct 
(e.g., ``book,'' ``plate''), in many cases, there are multiple
objects present with the same broad semantic label, making
it difficult for robots to identify the correct object. 
For example, it can be challenging for a system to identify 
``medication'' as a label. While human-human communication 
relies on identifying salient differentiating attributes 
(``the blue pill bottle,'' ``the prescription box''), reference to 
objects with differentiating attributes when communicating with 
humans and robots is a field of ongoing, fast-developing research.

Our objective is therefore not to mimic how humans refer to 
objects. We do not want to produce the type of ``natural'' labels 
that humans would use for communication. Instead, we will attempt 
to create simple and distinguishing labels that can be easily 
shared by humans and robots and that can be easily linked across 
language and visual modalities. As an example, consider a scenario 
where a user provides a vague command such as ``Fetch the 
medication from the shelf.'' While our approach may not be 
able to label the object as ``medication,'' the user can refer 
to a label provided by a real-time interface of the robot's 
view, ensuring that the robot retrieves the correct target 
object from the shelf. This is a clear advantage of our 
system, as it allows for modulation of the command and 
improves the accuracy of the robot's response in assisting 
individuals with their medication management.

The proposed models for addressing the RLBench benchmark 
consist of two main components: an input encoder that handles 
visual input from cameras and textual input from user instructions 
to produce encoded representations, and a transformer policy that
utilizes the encoded representations to predict the next robot 
action, typically described in terms of the robot's gripper's 
position, quaternion, and state. Our research builds upon these 
architectures by introducing an augmentation module to enhance 
textual and visual inputs. This augmentation aims to establish 
a clear reference between the object mentioned in the human 
command and a target object segmented or highlighted in the
robot's visual input.

To establish this reference, two key components are employed: 
a shared scene representation and unique synthetic labels
(see Fig. \ref{fig:demo} for an example). 
By using a shared scene representation, we provide a consistent 
understanding of the environment. Synthetic labels are introduced 
to replace potentially ambiguous references, allowing users to 
use specific identifiers such as ``object \#1'' to refer to target 
objects. To facilitate this, users are presented with an augmented 
real-time video feed of the robot's visual inputs, where each 
detected object is labeled with a synthetic label. These synthetic 
labels serve as a clear reference for an object named in the 
command.

When a user refers to a target object using its synthetic label, 
the augmentation module modifies the textual and visual inputs. 
The synthetic label in the textual input is replaced with a special 
token ``TARGET'', while the corresponding object in the visual input
is highlighted or marked. The specific method of marking the referred
object in the visual input, such as using bounding boxes or 
object segmentation boundaries, will be determined based on 
state-of-the-art visual object detection models. During training, 
we hypothesize that the multimodal encoder will learn to associate 
the special token ``TARGET'' with the marked object in the visual 
input, enabling the policy transformer to utilize this association 
for action prediction.

To ensure consistent object tracking across different camera angles 
and time steps, related work and state-of-the-art models in object 
detection and tracking are leveraged. We also consider the 
possibility of assigning unique synthetic labels to objects 
across different camera angles, allowing the augmentation of 
visual input for a specific camera angle containing the mentioned 
synthetic label. This exploration is essential as communication 
robustness improves when a single camera angle is sufficient for 
resolving object references.

In terms of data collection, we will adapt the crowdsourced 
dataset to include synthetic labels. This eliminates the need
for separately collecting instructions with synthetic 
labels. 

\section{Current Progress}

This section presents an overview of the current progress of our 
research, focusing on the successful setup of the RLBench environment
and the implementation of RLBench household tasks. Through the utilization
of this environment, we have identified specific tasks that lend themselves
well to the incorporation of language-based modulations, opening avenues 
for investigating the efficacy of our proposed approach.

To validate the feasibility of our approach, we initially concentrate our 
efforts on a single task as a proof of concept. Specifically, we selected 
the task ``stack the cups'' as our initial testbed. This task involves the
robot stacking two cups into a third cup, thereby establishing a tangible 
scenario for evaluating the effectiveness of language-based modulations in
enhancing robotic performance. Although we have identified many more complex
tasks that allow for modulation, we select a task with relatively short duration
and complexity to reduce computing and training times in this early
phase of the project.

In our initial experimentation, we introduced a modulation instruction in the 
form of ``stack the other cup first.'' This modulation aims to guide the robot's
actions by explicitly specifying the order in which the cups should be stacked. 
By incorporating language-based modulations into the task, our objective is to 
improve the robot's understanding of the task and enhance its ability
to reliably execute complex actions.

To assess the performance of our approach, we train a small-scale model with the aim of
solving the ``stack the cups'' task. The training process is in-line with models by \cite{james2020rlbench}. 
By focusing on a manageable task scale, we seek to evaluate the feasibility and 
effectiveness of our approach before scaling up to more intricate tasks, which
require vastly more computation.

Our current model remains relatively limited in terms of its capacity and capabilities, but
nonetheless allows us to collect images for all camera angles in order to implement the 
transparent approach. Moving forward, our immediate objective is to incorporate textual 
and image augmentation techniques to examine whether the proposed transparent approach further 
enhances the performance and reliablity of our model. 

At this critical juncture, feedback from the Geriatronics community becomes invaluable. 
We recognize the significance of soliciting input and insights from domain experts and
practitioners to refine our approach. The feedback obtained will guide us in iteratively
refining our methodology, ensuring that our proposed research aligns with the specific requirements
and challenges faced in the field of Geriatronics.

\section{Conclusion}

In conclusion, our research in language-based modulations for robotic systems 
demonstrated the potential of natural language instructions to enhance robot adaptability
and responsiveness. Through the use of the RLBench environment, we plan to validate the feasibility 
of integrating language-based modulations into robotic tasks, as exemplified by our proof-of-concept on 
the ``stacking cups'' task (Figure \ref{fig:demo}).

Collaboration with the Geriatronics community is vital in refining our approach and
addressing specific challenges in assistive robotics. With further research and 
development, we envision a future where assistive robots seamlessly integrate natural
language instructions, providing tailored support and enhancing the quality of life
for the elderly in Geriatronics applications.

\ifCLASSOPTIONcaptionsoff
  \newpage
\fi

\end{document}